\documentclass[manuscript,screen]{acmart}

\AtBeginDocument{%
  }

\setcopyright{acmlicensed}
\copyrightyear{20xx}
\acmYear{20xx}
\acmDOI{XXXXXXX.XXXXXXX}

\acmISBN{978-1-4503-XXXX-X/18/06}

\begin{document}
\title{Gaze estimation learning architecture as support to affective, social and cognitive studies in natural human-robot interaction}

\author{Maria Lombardi}
\email{maria.lombardi1@iit.it}
\orcid{0000-0001-5792-5889}
\authornote{Corresponding authors.}
\affiliation{%
  \institution{Humanoid Sensing and Perception, Istituto Italiano di Tecnologia}
  \city{Genoa}
  \country{Italy}
}

\author{Elisa Maiettini}
\email{elisa.maiettini@iit.it}
\affiliation{%
  \institution{Humanoid Sensing and Perception, Istituto Italiano di Tecnologia}
  \city{Genoa}
  \country{Italy}
}

\author{Agnieszka Wykowska}
\email{agnieszka.wykowska@iit.it}
\affiliation{%
  \institution{Social Cognition in Human-Robot Interaction, Istituto Italiano di Tecnologia}
  \city{Genoa}
  \country{Italy}
}

\author{Lorenzo Natale}
\email{lorenzo.natale@iit.it}
\authornotemark[1]
\affiliation{%
  \institution{Humanoid Sensing and Perception, Istituto Italiano di Tecnologia}
  \city{Genoa}
  \country{Italy}
}

\renewcommand{\shortauthors}{Lombardi et al.}

\begin{abstract}
Gaze is a crucial social cue in any interacting scenario and drives many mechanisms of social cognition (joint and shared attention, predicting human intention, coordination tasks). Gaze direction is an indication of social and emotional functions affecting the way the emotions are perceived. Evidence shows that embodied humanoid robots endowing social abilities can be seen as sophisticated stimuli to unravel many mechanisms of human social cognition while increasing engagement and ecological validity. In this context, building a robotic perception system to automatically estimate the human gaze only relying on robot's sensors is still demanding. Main goal of the paper is to propose a learning robotic architecture estimating the human gaze direction in table-top scenarios without any external hardware. Table-top tasks are largely used in many studies in experimental psychology because they are suitable to implement numerous scenarios allowing agents to collaborate while maintaining a face-to-face interaction. Such an architecture can provide a valuable support in studies where external hardware might represent an obstacle to spontaneous human behaviour, especially in environments less controlled than the laboratory (e.g., in clinical settings).
A novel dataset was also collected with the humanoid robot iCub, including images annotated from $24$ participants in different gaze conditions. 
\end{abstract}



\received{xx xxxx 20xx}
\received[revised]{xx xxxx 20xx}
\received[accepted]{xx xxxx 20xx}

\maketitle

\section{Introduction}
\label{sec:introduction}
Any human-human interaction is characterised by continuous nonverbal signalling that enriches, enforces and even drives the meaning of the verbal communication. The ability to perceive and understand the social information carried by such behavioural cues makes the communication more effective. Non-verbal communication includes facial expression, body language, gestures, gaze, tone and pitch of the voice, to cite few. Eye gaze is a strong social cue of non-verbal communication, and plays a crucial role in many mechanisms of social cognition, for example joint attention, engagement, signalling interest and intention, beating the time sequence of interpersonal interactions~\cite{hall2019,yin2022}. Gaze-following is one of the earliest and most essential ability developed by infants in engaging in social communication~\cite{mundy2007,csibra2011}. Three-months-old infants can follow an adult gaze orientation~\cite{hood1998,rochat2014} and, $3$ years old kids can accurately evaluate what individuals are looking at~\cite{doherty2006}. Further, in adulthood, this evolutionary cue triggers socio-cognitive processes, which allows individuals to engage in more social cooperation by inferring others’ behavioural intentions and mental states~\cite{dalmaso2020, cole2015}.

\subsection{Effect of gaze on human behaviour}
Given its importance, the effect of the gaze on human behaviour has been largely studied in the fields of neuroscience and psychology. For example, the so called \textit{gaze cueing effect} is a phenomenon in which human spatial attention shifts toward a location that has been gazed-at by another person. This causes enhanced processing of a stimulus that appears in that (gazed-at) location, relative to other locations, improving performance related to that stimulus (e.g., detection/discrimination reaction times)~\cite{friesen1998,frischen2007,mckay2021}.

Another phenomenon is the \textit{audience effect} which is a change that people have in their behaviour caused by the belief that someone else is watching them~\cite{canigueral2019a,canigueral2019b}. Interestingly, gaze has been also demonstrated to increase self-referential processing (see~\cite{conty2016} for a review). For example, Hietanen et al.~\cite{hietanen2008} investigated whether seeing another person’s mutual vs. averted gaze affected participants’ subjective evaluation of self-awareness. Results showed that mutual gaze resulted in an enhanced self-awareness rating by the participants, relative to averted gaze, most likely due to the presence of another person. This effect was observed only when participants were facing a real person, not when looking at a picture of a face.

Consequently, gaze measures such as gaze direction, looking time, anticipatory looking and gaze following, have become established measures in cognitive research~\cite{tafreshi2014,wilson2023}.

\subsection{Effect of gaze on human emotional state}
Going beyond the effect of the gaze on the human behaviour, gaze direction is also indication of social and emotional functions. People tend to look at things they like and avoid things they do not like (approach-avoidance)~\cite{shimojo2003,bayliss2007}. 
Previous research has found that when gaze direction matches the underlying behavioural intent (approach-avoidance) communicated by an emotional expression, the perception of that emotion is enhanced or facilitated~\cite{adams2003,adams2005}. For example, angry people often stare into the eyes of the person with whom they are trying to quarrel or fight, and timid people who fear others may avert their eyes and look away.

Moreover, gaze can have a significant influence on emotional expression perception. 
The study in~\cite{liang2021} found that gaze direction systematically influenced the perceived emotion disposition conveyed by a neutral face. Specifically, a direct gaze was attributed to a more joyful disposition, whereas an averted gaze was attributed to an angrier or more fearful disposition.

The close relation between the gaze direction and the emotional valence has been studied in~\cite{lombardi2023}. Specifically, the focus was on whether the gaze direction (mutual vs. averted), the type of emotion (happy vs. sad face) or their combined effect, being the outcomes of participants’ actions, modulated implicit Sense of Agency. Results showed that communicative gaze (i.e., both mutual and averted gaze) modulated SoA.

\subsection{Robots as tools to study human cognition}
Wykowska in \cite{wykowska2020,wykowska2021} highlights the role that robots can play as tools for examining human cognition, specifically in understanding mechanisms such as joint attention and sense of agency. By using robots (and specifically humanoid robots) as sophisticated stimuli it is possible to generalise the findings observed in classical experiments with 2D stimuli on a screen to more interactive and naturalistic scenarios. Furthermore, through their embodied physical presence, the use of humanoid robots in an experimental scenario provides higher ecological validity than 2D screen-based stimuli and better experimental control than human-human interaction.

In this framework, having a robotic perception system that is able to automatically estimate the human focus of attention through estimating gaze direction is still an open challenge. Indeed, it is so common seeing recent studies aiming at investing human-robot interaction where the robot still acts in a pre-programmed way without any feedback from the environments and the human. To meet the requirements of the field, the embodied robotic system needs to be self-contained without relying on external hardware.

Specifically, the proposed architecture has multiple functions: 1) estimating the human gaze direction as vector in a table top interactive scenario; 2) detecting if the human is looking at robot's body parts signalling human attention towards the robot as physical presence (e.g. robot's shoulders, arms or chest) and 3) detecting if the human is establishing events of mutual/averted gaze with the robot. 

Mutual gaze (or eye contact) is a particular case of the gaze estimation problem and occurs when two entities direct their gaze at each other's eyes at the same time~\cite{hietanen2018,jongerius2020}. In addition, our approach aims at reducing the amount of sensors and hardware required to estimate the gaze of the human partner, for example mobile or static eye-tracker, external cameras. The proposed architecture, indeed, relies only on the image frames captured by the robot's eye-like cameras in order to have an interaction as natural as possible with the human. Such a system provides a valuable support in experimental psychology, especially in all the studies where the external hardware is seen as an obstacle (for example, in clinical settings where the environment is highly dynamic and less controlled than the laboratory) or simply because it affects the spontaneity of the human behaviour.

\section{Related work and Contributions}
\label{sec:related_work}
The problem of endowing robot with the social ability to estimate the human gaze in terms of 2D or 3D gaze vector has been largely studied in the literature.\\
Several solutions leveraged convolutional neural networks (CNN) to compute a feature vector from RGB images and produce the prediction of the gaze vector as output. For example,~\cite{zhang2017} proposed a full-face CNN architecture for gaze estimation that takes the full face image as input and directly regresses to 2D or 3D gaze estimates. The CNN architecture was provided with a spatial weights mechanism to efficiently encode information from different regions of the full face, showing that such a mechanism both facilitates the learning process and makes the algorithm robust to variations in light conditions.

A convolutional architecture was also used in~\cite{athavale2022} to estimate the 2D coordinates $(x,y)$ of the gaze with the novelty of extracting features from the image of only one eye. Authors assessed that such an approach is crucial especially in real-world conditions where the human face can be partially obscured (traditional CNN trained on the whole face likely fail).
While previous work claimed that only one eye is enough to solve the problem of gaze estimation, a completely different approach was followed in~\cite{cheng2020} where the core strategy lied in the idea of exploiting the asymmetry and the difference between left and right eye to predict a combined 3D gaze vector from the RGB image. That was done combining a face-based asymmetric regression network (FAR-NET) and an evaluation network (E-Net) taking as input both the face image and the crops of the two eyes.

A learning approach based on two fully-convolutional pathways was, instead, proposed in~\cite{chong2018}. The neural network model takes three inputs: the whole image (processed by the scene pathway), a crop of the subject's face (processed by the face pathway) and the location of the face inside the image (combined with the output of the two pathways in a resulting fully connected layer). Given the input, the model estimates both the gaze vector in terms of yaw and pitch degrees, and the attention saliency heatmap.
A more recent approach combined the convolutional layers of ResNet18 with a trasformer architecture~\cite{cheng2022}. The architecture first employs a CNN backbone to extract local feature maps and then estimate 2D gaze from the feature maps with a transformer encoder.

Although gaze estimation and human attention have been extensively studied, very few studies exist on the detection of mutual/averted gaze between two interacting partners. To fill this lack in the literature, in~\cite{lombardi2022} the authors proposed a learning-based architecture able to automatically detect when the human is establishing mutual gaze with the humanoid robot iCub in real-time scenarios. The proposed approach was validated both computationally and in a real experimental setup. Precisely a variant of Posner paradigm~\cite{posner1980} adapted for human-robot interaction~\cite{kompatsiari2018} was used as it is one of the most used paradigm in the psychological studies regarding the gaze.

The learning architecture proposed in this paper exploits and extends the approach proposed in~\cite{lombardi2022} in order to detect not only mutual/averted gaze events but to make the humanoid robot iCub also aware of the human attention when they are looking elsewhere (e.g., robot's body parts and workspace defined in a table-top scenario).

The main contributions of this work are as follows:
\begin{itemize}
    \item \textit{Dataset collection, both for mutual/averted gaze detection and for gaze vector prediction in frontal human-robot interaction.} The dataset was collected to be general enough and, thus, suitable in several different experimental scenarios in the context of table-top tasks. To the best of our knowledge, it is the first dataset involving not only the annotation to predict the gaze direction as vector but also annotations for mutual/averted gaze and if the human is looking at robot's body parts.
    \item \textit{Designing and training of the learning gaze estimation architecture.} We propose a learning architecture, trained on the aforementioned dataset, to estimate the human attention while they are in front of the humanoid robot. Our approach relies only on the image captured by robot cameras and, thus, it minimises the amount of external hardware required to fulfil the estimation task. The system is validated and the performance reported both end-to-end and for each module composing the learning architecture.
    \item \textit{Comparison of the proposed architecture against the state-of-the-art.} To show its effectiveness, we compare our method with two solutions existing in literature and proposed in~\cite{3DGazeNet} and~\cite{VTD} showing better performance. More details on the comparison are reported in Section~\ref{sec:comparison_soa}.
\end{itemize}

\section{Data collection}
Data collection was conducted mainly targeting human-robot interactions in table-top scenarios (which is recurring in many experimental psychology setups in the literature~\cite{lombardi2023,antonj2023,hoffmann2021,kelley2021,marchesi2020}). It consists in three different sessions: 1) human establishing mutual/averted gaze with the robot; 2) human looking at different points on the table; 3) human looking at different parts of robot's body.

\subsection{Participants}
A total of $24$ participants were recruited for the data collection (mean age $= 29.54 \pm 3.14$, $14$ females and $1$ non-binary). All participants had a normal or corrected-to-normal vision ($6$ participants out of $24$ wore glasses). 
The data collection was conducted at the Istituto Italiano di Tecnologia in Genoa and it was approved by the local ethical committee (Comitato Etico Regione Liguria), and carried on in accordance with the ethical standards laid down in the 2013 Declaration of Helsinki. Before the data collection, all participants gave written informed consent and they were all debriefed about the purpose of the task.

\begin{figure*}[!t]
    \centering
    \includegraphics[width=0.7\textwidth]{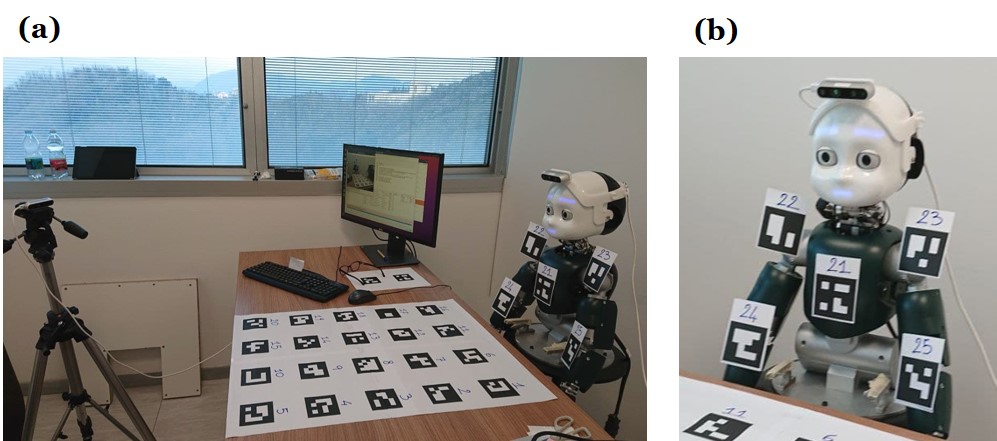}
    \caption{\textbf{Setup.} \textbf{(a)} Overall setup consisting in an aruco board, iCub robot with mounted a Realsense camera on its head and an external camera on a tripod. \textbf{(b)} Focus on iCub robot with aruco markers attached on its body.}
    \label{fig:setup}
\end{figure*}

\subsection{Setup}
Data was acquired using the humanoid robot iCub~\cite{icub}, embedding two Dragonfly2 cameras\footnote{http://wiki.icub.org/images/c/c9/POINTGREY\_-\_Dragonfly2.pdf} in the left and right eyes; only the right eye-camera was used with the frame resolution set to $640\times480$ pixels, but the left-eye camera could be used equivalently. In order to have also higher quality images for the training phase of the proposed architecture, the same frames were also collected with the Intel RealSense depth camera D435\footnote{https://www.intelrealsense.com/depth-camera-d435/} mounted on the iCub's head through a 3D printed headset.
Since the data collection does not require a full-body robot, an half-body version of the iCub robot was placed at the opposite side of the desk, at a distance of around one meter from the participant, and mounted on a support of $82$ cm high to ensure that its eyes were aligned with participants’ eyes.

In order to automatically acquire the ground truth of the gazed points 3D coordinates when the human is looking at the table, a $5\times4$ Aruco board\footnote{https://docs.opencv.org/4.x/db/da9/tutorial\_aruco\_board\_detec-tion.html} was placed on the table between the participant and the iCub in order to cover all the workspace of interest. Extensively used in pose estimation, an Aruco board is a set of uniquely identified squared binary matrix (called markers) that acts like a single marker in the sense that it provides a single pose for the camera. The id of the marker corresponds to a precise binary codification of the matrix.
In our case, each marker of the aruco board was chosen to be a $7\times7$ square, $7$ cm distant from each other on all sides. The whole board measures $63\times49$ cm with a total resolution of $7449\times5787$ pixels. The main advantage of using an aruco board is that, while only one marker is sufficient for localisation of the whole board, having multiple markers is beneficial when part of the board is outside the field of view to increase robustness to occlusions or misdetections.
Furthermore, $5$ aruco markers were placed on the iCub's body parts, precisely, on the shoulders, on the forearms and on the chest, to estimate the ground truth of the 3D space coordinates for the human looking at the iCub robot. To complete the recording setup, a further external Intel Realsense depth camera 415\footnote{https://www.intelrealsense.com/depth-camera-d415/} was positioned on a tripod in order to capture the setup by a third-person view.

The recording setup is shown in Figure~\ref{fig:setup}. In line with the aim of this study, claimed in Section~\ref{sec:introduction} - i.e. avoiding all the external hardware that can alter the results - we underline that both Realsense cameras (i.e., the external one and the one mounted on the iCub's head) were used only for acquiring data for the training phase of the proposed method. In the deployment phase, the system was tested and the performance evaluated only using images provided by the eye-camera embedded in the iCub's right eye.

The iCub robot was connected over a local network to a workstation acting as a server and to another client laptop equipped with an external GPU (NVIDIA GeForce GTX 1080). Specifically, the workstation, was used to launch and control the experiments of the clients, whereas the client laptop with the GPU was effectively used to process and store the acquired data. The middleware YARP\footnote{https://www.yarp.it/latest/} (Yet Another Robot Platform)~\cite{yarp} was used to coordinate and integrate the different modules of the setup (e.g. realsense cameras, iCub's camera, iCub’s controller, data dumper, python code modules) in order to let them communicate with each other in a peer-to-peer manner.

\subsection{Task}
At the beginning of the data collection, participants were asked to sit in front of the iCub robot. A keyboard was given to the participants, allowing them to acquire the RGB frame pressing the space bar when they were ready. The details for each session are reported in the following paragraphs.\\

\begin{figure*}[!t]
    \centering
    \includegraphics[width=0.7\textwidth]{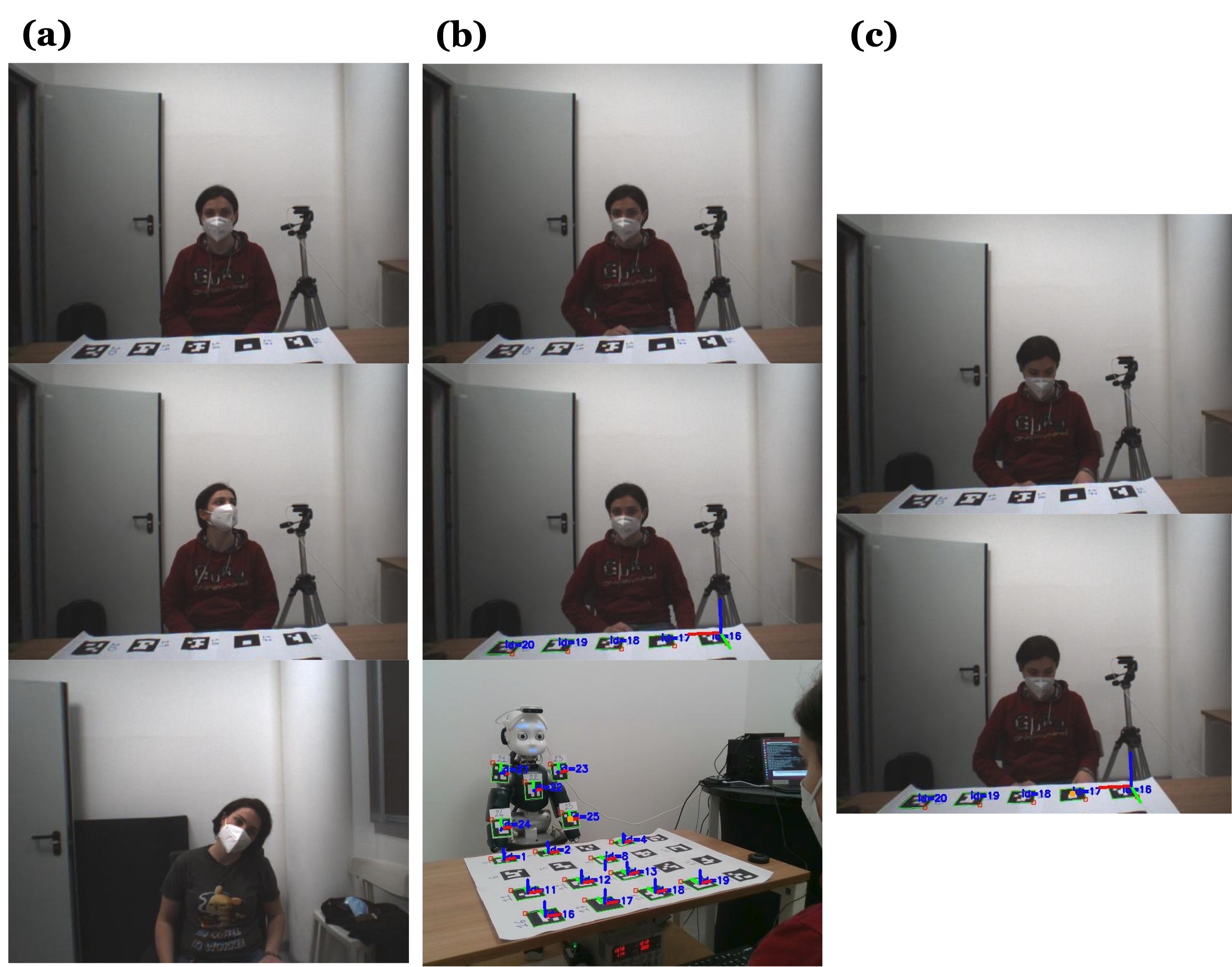}
    \caption{\textbf{Dataset collection.} \textbf{(a)} Sample frames acquired during the mutual/averted gaze session of the data collection. The participant was asked to look at the robot face in different head position (e.g. frontally and rotating the head) and do not look at the robot (averted gaze). \textbf{(b)} Samples frames acquired while the participant was looking at the robot's body part (e.g. left forearm). Different views were taken of the human (frontally and back) in order to show the detection of the aruco markers both from the iCub's eye and external Realsense camera. \textbf{(c)} Samples frame acquired while the participant was looking at the aruco board placed on the table. The detection of the aruco board in shown also in this scenario.}
    \label{fig:dataset}
\end{figure*}

\textbf{Mutual/averted gaze}. Each participant was instructed firstly to naturally look at iCub's eyes and then look at any other point they preferred in the room in order to acquire frames both in eye-contact and no eye-contact conditions. In the eye-contact condition, each participant was also asked to keep their gaze on iCub's eyes but rotating laterally first their torso and then their head. For each position, a frame was captured both by the iCub's right camera and the RealSense mounted on the headset, at the time when participant pressed the space bar on the provided keyboard (Figure~\ref{fig:dataset}a).\\

\textbf{Gazing at iCub's body parts}. Each participant was asked to naturally look at each of the five aruco markers placed on iCub's body upon the verbal instruction given by the experimenter (e.g., ``look at the marker $23$'') and to press the space bar when they were ready. Frames were also captured instructing the participant to keep their gaze on each marker but slightly rotating the head up, down, left and right. Several frames were collected for each position (Figure~\ref{fig:dataset}b).\\

\textbf{Gazing at the workspace}. Each participant was asked to naturally look at a specific aruco marker on the board upon a command like, e.g., ``look at the marker $3$'', given by the experimenter and press the bar space to capture the frame when they were ready. The order of the markers was decided randomly by the experimenter. Several frames were collected for each aruco marker (Figure~\ref{fig:dataset}c).

The final datasets consist of $1445$ frames for each camera (iCub's camera and Realsense mounted on the head).

\subsection{Data annotation}
\label{sec:data_annotation}
\begin{figure*}[!t]
    \centering
    \includegraphics[width=\textwidth]{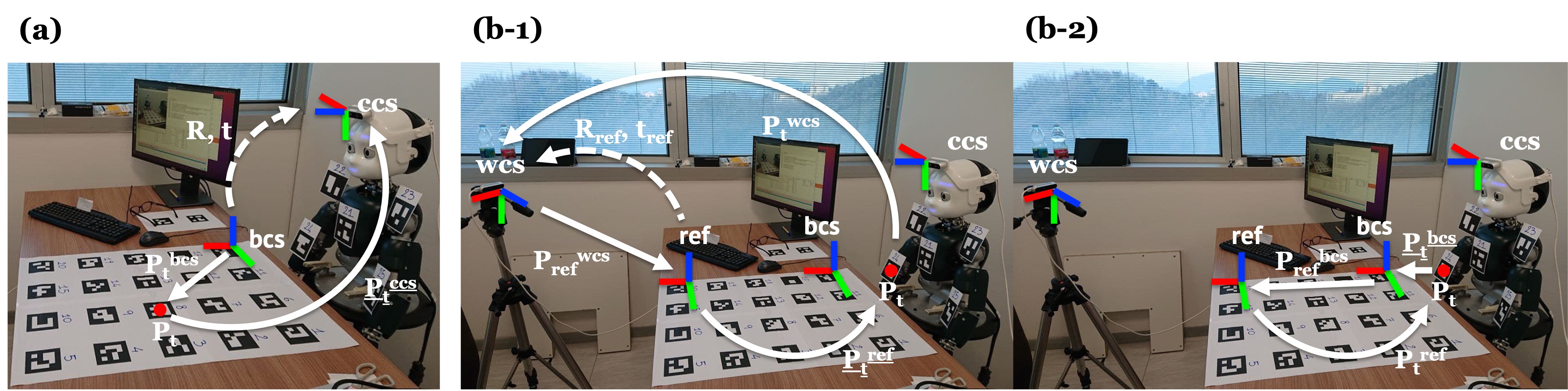}
    \caption{\textbf{Dataset annotation.} Pictorial representations of the data annotation process reporting the geometric transformations to have the target point in the camera coordinate system (CCS) passing through different coordinate systems (WCS, BCS and REF) used as support. \textbf{(a)} Data annotation for the scenario ``Gazing at workspace''. \textbf{(b)} Data annotation for the scenario ``Gazing at iCub's body parts''. The target point $P_t$ is marked in red, the dashed white lines indicate the transformation between the different coordinate systems whereas the solid white line indicate the point in the corresponding system. Finally, the resulting point of interest is underlined.}
    \label{fig:geometric}
\end{figure*}

Data were annotated in different ways according to the aim of each session: 1) mutual/averted gaze; 2) human looking at the robot's body parts; 3) human looking at the workspace (e.g., the table). See Figure~\ref{fig:dataset} for a pictorial representation of the three considered scenarios and Figure~\ref{fig:geometric} to visually support the data annotation process described in what follows.\\

\textbf{Mutual/averted gaze.} In this scenario (Figure~\ref{fig:dataset}a), the YARP data dumper module\footnote{https://www.yarp.it/latest/group\_\_yarpdatadumper.html} was used to store only the RGB frames acquired from both cameras when the participant released the space bar on the keyboard. The corresponding annotation ``eye-contact/no eye-contact'' for each batch of frames was written manually by the experimenter.\\

\textbf{Gazing at the workspace.} In this case, additional information was stored with respect to the previous one (Figure~\ref{fig:dataset}c). Specifically, for each gazed aruco marker on the board (upon the experimenter's instruction), the data stored at the space bar release trigger were:
\begin{itemize}
    \item RGB frames from the iCub's camera and Realsense camera on the iCub's headset;
    \item Depth matrix from the Realsense camera mounted on the robot's head;
    \item Human face centroid $P_o$: OpenPose~\cite{openpose} was used as pose estimator to evaluate the image coordinates $(u,v)$ of the centroid of the face landmarks, computed as the mean coordinates of all face keypoints. Knowing the intrinsic camera parameters, the point $P_o$ (in pixel) can be converted and saved also as a point $(x,y,z)$ in camera coordinate system (CCS) through the well-known formulas:
    $x = \frac{u - c_x}{f} z$; $y = \frac{v - c_y}{f} z$; $z = d$, where $d$ is the depth acquired from the Realsense camera, corresponding to the pixel $(u,v)$, $f$ is the focal distance and $(c_x, c_y)$ are the coordinates of the principal point of the camera. Since the iCub's camera does not provide any depth, in this case $d$ was kept constant and estimated around one meter (this is reasonable because the participant was sitting around one meter far from the robot).
    \item Gaze target point $P_t$: as mentioned above, the main advantage of using an aruco board with OpenCV libraries\footnote{https://docs.opencv.org/4.x/db/da9/tutorial\_aruco\_board \_detection.html} is that we can easily know the position of the board origin in the camera coordinate system in terms of the rotation vector $r$ and traslation vector $t$ using only the intrinsic camera parameters. In detail, let $P_t^{bcs}$ be the position of a generic marker in the board coordinate system BCS (in our case the origin is centered in the marker id $16$ - and with the z-axis pointing out from the plane). The point $P_t^{bcs}$ can be easily reconstructed in BCS directly from the board layout defined a-priori. The corresponding point $P_t$ in CCS can be defined as: 
    \begin{equation}
        P_t = \begin{bmatrix}
            R & t\\
            0 & 1
            \end{bmatrix} P_t^{bcs}
    \end{equation}
    where $t$ is the traslation vector and $R$ is the rotation matrix derived by the rotation vector $r$ using the Rodrigues' rotation formula (Figure~\ref{fig:geometric}a).\\
    Similar to what has been done before, the 3D point of the aruco in the CCS can be converted in the 2D space of the image plane following the well-know relations: $u = f \frac{x}{z} + c_x$ and $v = f \frac{y}{z} + c_y$, where $f$ is the focal distance and $(c_x, c_y)$ are the coordinates of the principal point of the camera.
    \item Aruco board rotation and translation vector used as extrinsic camera parameters with respect to the CCS.\\
\end{itemize}

\textbf{Gazing at iCub's body parts}. In this last scenario (Figure~\ref{fig:dataset}b), the robot body parts of interest were marked with aruco markers in order to annotate their positions in the camera reference frame. Since such aruco markers were not visible from iCub's camera, a second Realsense camera mounted on a tripod was used to take a third point of view of the scene. Specifically, for each aruco marker gazed on the robot body parts (upon the experimenter's instruction), the data stored on the release of the space bar were:
\begin{itemize}
    \item RGB frame from the iCub's camera, Realsense camera mounted on the robot headset and the external Realsense camera mounted on a tripod;
    \item Depth matrix from the Realsense camera mounted on the robot's head;
    \item Human face centroid $P_o$ was evaluated as before;
    \item Gaze target point $P_t$: exploiting the aruco board properties and the known intrisic parameters of both cameras (external and the first point of view camera) the target point $P_t$ in CCS was evaluated as follows. Let $P_{ref}$ be the point of the aruco marker with the highest id visible from the external camera and let define $P_{ref}^{wcs}$ as the reference point in world coordinate system (WCS), easily computed through the intrinsic parameter of the external camera. We can evaluate the target point $P_t^{ref}$ in the reference system (centered in the point used as reference) as (Figure~\ref{fig:geometric}b-1):
    \begin{equation}
        P_t^{ref} = R_{ref}^{T} \left(P_t^{wcs} - P_{ref}^{wcs} \right)
    \end{equation}
    where $R_{ref}^{T}$ is the transpose of the rotation matrix of the reference marker respect to the WCS, $P_t^{wcs}$ and $P_{ref}^{wcs}$ are respectively the target point and the reference point in WCS. From this, we can easily compute the target point $P_t^{bcs}$ in board coordinate system (BCS) (Figure~\ref{fig:geometric}b-2):
    \begin{equation}
        P_t^{bcs} = P_{ref}^{bcs} + P_t^{ref}
    \end{equation}
    where $P_{ref}^{bcs}$ is the reference point in BCS computed as offset from the origin through the aruco board layout.
    Having the target point $P_t^{bcs}$ expressed in the board coordinate system, finally $P_t$ in CCS is (Figure~\ref{fig:geometric}a):
    \begin{equation}
        P_t = \begin{bmatrix}
            R_{b} & t_{b}\\
            0 & 1
            \end{bmatrix} P_t^{bcs}
    \end{equation}
     where $t_{b}$ is the board traslation vector and $R_b$ is the board rotation matrix derived by the rotation vector $r$ using the Rodrigues' rotation formula respect to the CCS.
    \item Aruco board rotation and translation vector used as extrinsic camera parameters with respect to the camera coordinate system and world coordinate system.
\end{itemize}

\section{Gaze estimator architecture}
\begin{figure*}[!t]
    \centering
    \includegraphics[width=0.9\textwidth]{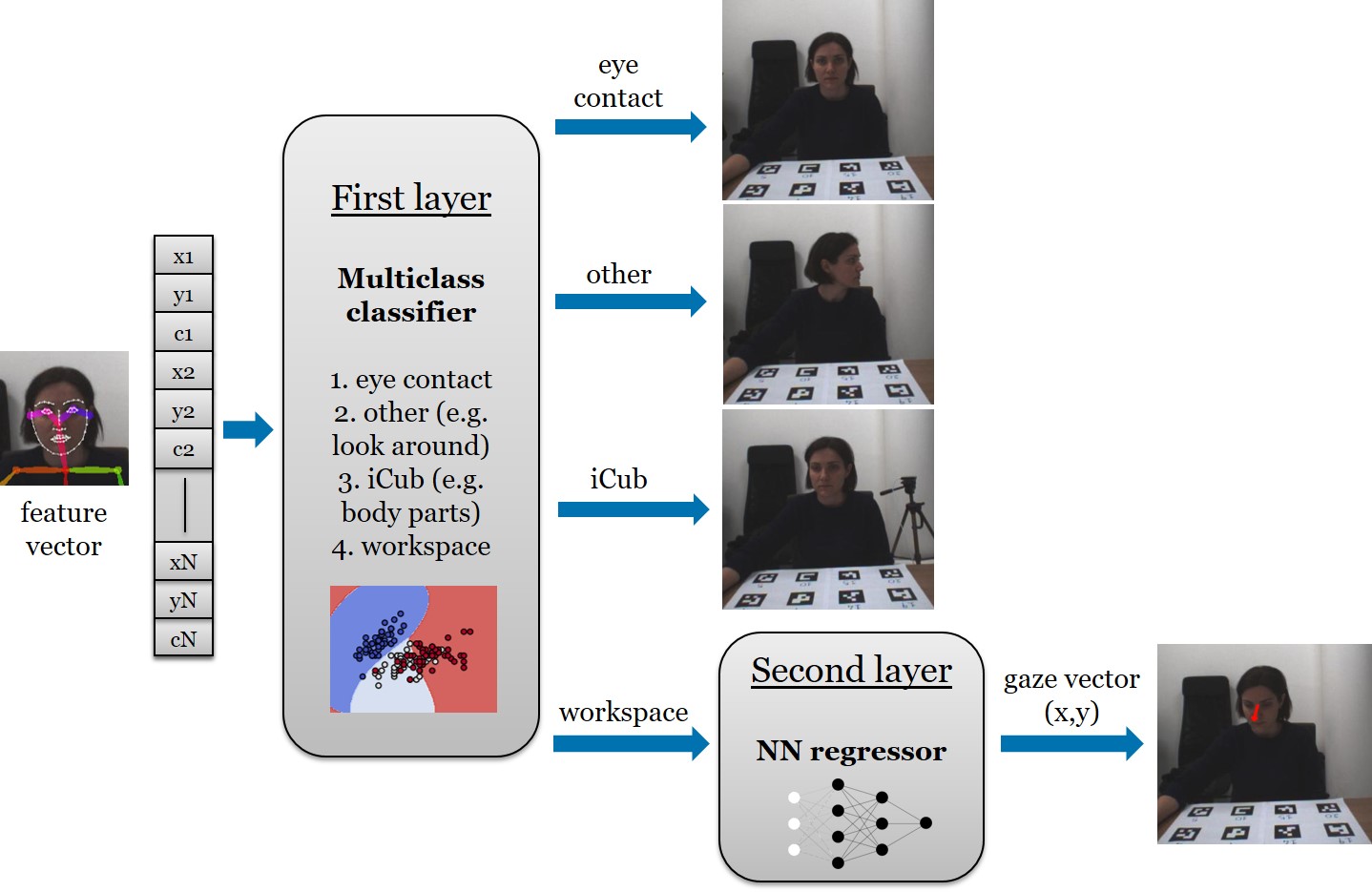}
    \caption{\textbf{Learning architecture.} The feature vector extracted by OpenPose is used as input to the multiclass classifier whose output is the pair $\left(r,c\right)$, where $r\in$ (eye\_contact, other, iCub, workspace) and $c$ is the confidence level. Only if the multiclass classifier has the class ``workspace'' as output, the feature vector is forwarded to the second layer of the architecture and so to the gaze regressor. The final output is the vector $\left(x,y\right)$ representing the 2D gaze vector with the corresponding confidence level.}
    \label{fig:architecture}
\end{figure*}

Once the dataset was collected as described in the previous Section, the multi-pose estimator OpenPose~\cite{openpose} was used to extract the feature vector from each frame image. Specifically, for each image, OpenPose produces a vector representing 2D locations $\left(x,y\right)$ of anatomical keypoints for each person in the scene with the corresponding confidence level, $k$. In our work, a subset of $19$ keypoints are considered (out of the $135$ predicted by OpenPose). Specifically, they are $8$ points for each eye, $2$ points for the ears and $1$ point for the nose, resulting in a feature vector of $57$ elements (for each point the triplet $\left(x,y,k\right)$ is taken). The feature vector is then centred with respect to the head centroid and normalised on the farthest point from it. The use of the face keypoints as feature vector has the main advantage to be robust to changes in the frame, such as background and light conditions. The resulting feature vector is used as input to a two-layer architecture mainly composed by: 1) multi-class classifier and 2) gaze regressor (see Figure~\ref{fig:architecture}).

\subsection{Multi-class classifier}
\label{sec:multiclass_classifier}
The multi-class classifier takes the feature vector as input and produces in output one of the following four classes: \textit{eye\_contact} (when mutual gaze is established between the human participant and the robot), \textit{icub} (when the human looks at the robot body parts), \textit{workspace} (the human looks down at the table defined as workspace) and \textit{other} (the human is looking anywhere else in the room). Each prediction is always given together with the corresponding confidence level. Support Vector Classifier (SVC) with RBF kernel was chosen to address this classification task.

We compared the SVC with a random forest classifier and a feedforward neural network; the former was chosen because it reported the best performance in terms of accuracy and F1-score (for a detailed comparison, see \textit{Supplementary Material}).
The multiclass support was handled according to the ``ovr'' scheme (one-vs-rest) while the hyperparameters of the SVC model were selected using an exhaustive search over a grid parameters and optimised by a $5$-fold cross-validation~\cite{scikit-learn}.

The part of dataset labelled with the class \textit{eye\_contact} and \textit{icub} was augmented in order to be robust to the degenerative case in which OpenPose fails to detect the eyes' boundaries and the pupils. To simulate such a scenario, we augmented the training set with additional samples obtained by setting to zero the part of the feature vector corresponding to the coordinates of those keypoints; the rest was left unchanged. The samples labelled with the class \textit{eye\_contact} was further augmented by geometrically rotating the feature vector to the left and right of a certain angle around the face centroid. Specifically, the keypoints were rotated to the left and to the right by an angle $\alpha \in \{15^{\circ}, 30^{\circ}, 45^{\circ}, 60^{\circ}\}$ taking the  $\{5\%, 10\%, 10\%, 5\%\}$ of the data, respectively. The dataset corresponding to the class \textit{workspace} and \textit{other} was left unchanged. We handled the resulting unbalanced dataset properly weighting each class. Such weights were chosen inversely proportional to class frequencies in the input data~\cite{scikit-learn}.

The classifier was trained (as described above) both using the dataset collected with the RealSense and with iCub’s eye. As test, only images taken by the eye camera were used, because the goal was to evaluate the performance of the architecture when only images from the cameras on board the robot were available. For the full comparison between the two datasets, see \textit{Supplementary Material}.

Only the data samples classified as \textit{workspace} are then forwarded to the second layer of the architecture, namely the gaze regressor.

\subsection{Gaze regressor}
\label{sec:gaze_regressor}
The gaze regressor takes the normalised feature vector forwarded from the previous layer as input. Each pair coordinate-confidence $(x_i, c_i)$ and $(y_i, c_i)$ for each face keypoint $i$ of the feature vector is then forwarded to a layer made up of $38$ Confidence Gate Unit (one for each two pairs of the $19$ used keypoints). The Confidence Gate Unit has been initially proposed in~\cite{dias2020} and here adapted to address our specific case. Specifically, the Confidence Gated Unit (CGU) weights each coordinates for the corresponding confidence level and is mainly composed of two internal units: i) a ReLU unit acting on the single coordinate $x_i$ (or $y_i$) of the input feature and ii) a sigmoid function acting as a gate on the confidence level $c_i$. The two results $CGU(x_i)$ and $CGU(c_i)$ are then multiplied with each other and concatenated with the output of all the other CGUs. The resulting vector is then used as input to other two fully connected layers of $19$ hidden nodes each. The output layer has $3$ nodes corresponding to the triplet $(g_c, g_y, \sigma)$ where $(g_x, g_y)$ are the coordinates in pixels of the estimated gaze vector and $\sigma$ is the corresponding confidence level.

\subsubsection{Training details} 
Having the dataset described in Section~\ref{sec:data_annotation}, the gaze versor used as ground truth was calculated as the direction of the line passing through the head centroid and the gaze target point on the aruco board. Specifically, the gaze vector calculated as distance between the head centroid and the target point on the aruco board was first normalised as versor $10$cm long in camera coordinate system and then projected in the image plane (the vector's length of $10$cm was arbitrarily chosen). So doing, the module of the bidimensional vector in the image plane models the z-axis of the original 3D gaze vector. Similarly to what described in Section~\ref{sec:multiclass_classifier} for the multi-class classifier, the workspace dataset was also augmented to include $40\%$ of training samples obtained by setting to zero the part of the feature vector corresponding to the keypoints of the eyes.

The neural network hyper-parameters were tuned empirically. A normal kernel initializer was used for the two fully connected layers while the parameters for the CGU layer were set all to one. As regularisation, L2 penalty of $10^{-4}$ was applied to the parameters of the two fully connected layers, L2 penalty of $10^{-3}$ was applied for the CGU layer while the input and output layers had no regularisation. Furthermore, the training was done over $100$ epochs with batch size of $400$ samples and using the Adam optimiser (learning rate set to $0.05$ with a decay of $0.9$ and ReLU as activation function)~\cite{kinga2015}. Finally root mean square error (RMSE) was used as loss function between the ground truth and the prediction.

The regressor was trained both using the dataset collected with the RealSense and with iCub’s eye (as done for the multiclass classifier) while, also in this case, the test set was obtained using only images from the camera on the iCub's eye. For the full comparison between the two datasets, see \textit{Supplementary Material}.

\section{Results}
\label{sec:results}

\subsection{Evaluation of the multi-class classifier and the gaze regressor on the collected test set}
For the training, the dataset was split into two subsets taking $19$ out of $24$ participants for the training set and the others $5$ participants for the test set. The dataset was split $k = 5$ times in order to average the performance over different participants subsets. Different models were trained using the training set from the Realsense and iCub's camera individually or jointly. Regardless of the training set the performance was evaluated on the test set collected by the iCub's camera.

The multiclass classifier was evaluated in terms of accuracy, precision, recall and F1-score. The highest performance was reached combining the dataset collected by iCub with that one collected by the Realsense resulting in an accuracy value of $0.87 \pm 0.03$ (see Table ~\ref{tab:multiclass_classifier} for all the results).
The gaze regressor, instead, was evaluated in terms of RMSE between the ground truth (acquired though the aruco markers as explained in Section~\ref{sec:data_annotation}) and the predicted 2D gaze vector. Precisely, the gaze regressor reached an error in pixels of $5.27 \pm 0.88$ with the trainset acquired from iCub's camera, $5.43 \pm 0.79$ with that one acquired from the Realsense and $5.26 \pm 0.74$ combining both train sets (note that the resolution of the RGB frames in input is $640\times480$). Furthermore, the small standard variation indicates that the error reported by the model has not significant variation over the different splits.

\begin{table}[h!]
    \renewcommand{\arraystretch}{1.2}
    \caption{\textbf{Performance of the multiclass classifier on iCub's test set.} Mean and standard deviation of accuracy, precision, recall and F1 score are reported for the SVC algorithm trained with the data acquired from the iCub's camera and/or from the Intel RealSense camera and tested on the data collected only from the iCub's eye.}
    \centering
    \begin{tabular}{|c|c|c|c|}
         \cline{2-4}
         \multicolumn{1}{c|}{} & \multicolumn{3}{|c|}{\textit{iCub Test set}}  \\
         \hline
         \textit{Train set} & \textit{iCub} & \textit{Realsense} & \textit{iCub + Realsense} \\
         \hline
         Accuracy & $0.85 \pm 0.03$ & $0.76 \pm 0.05$ & $\mathbf{0.87 \pm 0.03}$ \\
         \hline
         Precision & $0.75 \pm 0.05$ & $0.65 \pm 0.04$ & $\mathbf{0.79 \pm 0.06}$ \\
         \hline
         Recall & $0.72 \pm 0.06$ & $0.69 \pm 0.04$ & $\mathbf{0.75 \pm 0.05}$ \\
         \hline
         F1 score & $0.73 \pm 0.05$ & $0.66 \pm 0.04$ & $\mathbf{0.75 \pm 0.05}$ \\
         \hline
    \end{tabular}
    \label{tab:multiclass_classifier}
\end{table}

\subsection{End-to-end evaluation of the pipeline}
In this Section, the performance of the whole architecture are evaluated end-to-end. Specifically, for each of the five splits of the dataset, the pipeline was built using the corresponding models trained on that split and the performance evaluated on the corresponding test set. Also in this case, the models used for the multiclass classifier and the gaze regressor are those trained both with the data from iCub's camera, Realsense camera and the dataset coming from both of them. The results are reported in Table~\ref{tab:end-to-end} as mean and standard deviation over the dataset splits. The used metrics are as follows: 
\begin{itemize}
    \item fraction of data classified correctly as \textit{workspace}, representing the amount of data forwarded to the second layer to the gaze regressor (Figure~\ref{fig:architecture});
    \item root mean square error between the ground truth and the predicted 2D gaze vector (output of the architecture);
    \item angular error in degrees between the ground truth and the reconstructed 3D gaze vector in camera coordinate system.
\end{itemize}

\begin{figure*}[!t]
    \centering
    \includegraphics[width=\textwidth]{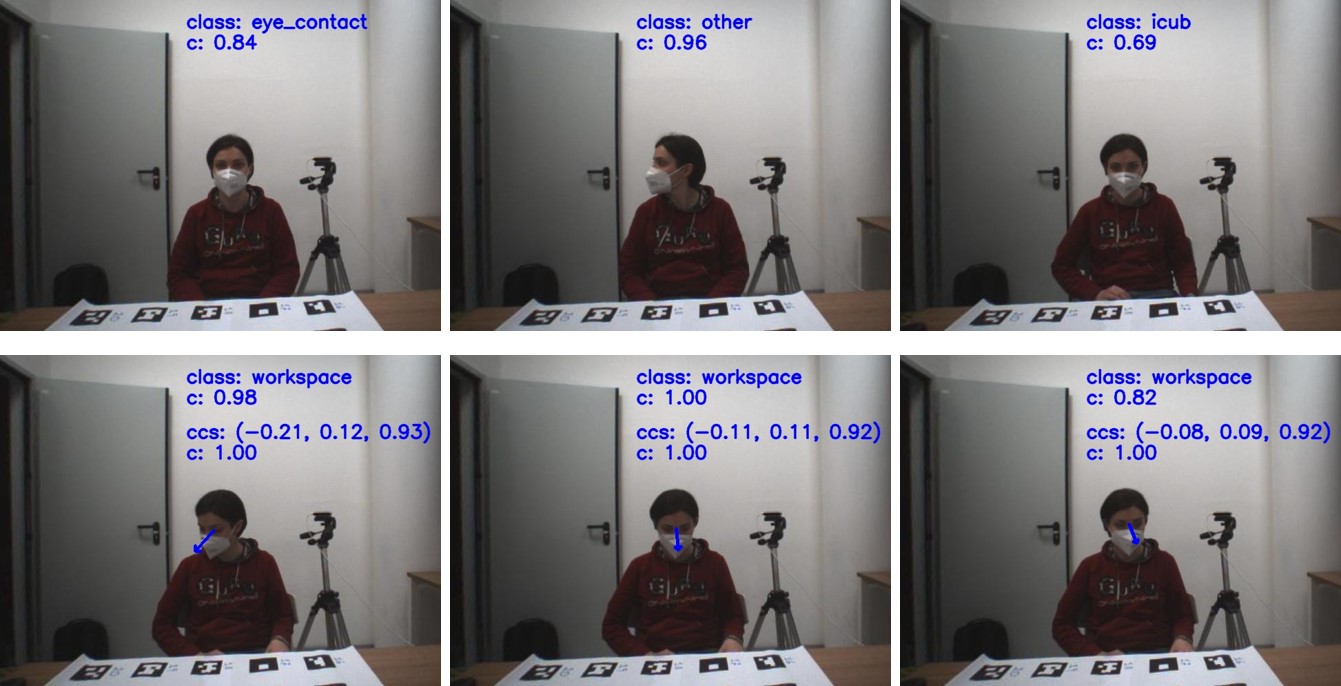}
    \caption{\textbf{Architecture's output.} Sample frames taken from the test set to show the output of the learning pipeline in the different considered scenarios: human in mutual gaze, human in averted gaze, human looking at robot's parts and human looking at the workspace. For each frame the multi-class classifier's output is reported with the corresponding confidence level. For the frames classified as workspace, also the predicted gaze vector and the reconstructed coordinates of the gaze in camera frame reference are reported.}
    \label{fig:results_testset}
\end{figure*}

A geometric approach was used to reconstruct the 3D gaze vector from the bidimensional gaze predicted from the architecture (see Figure~\ref{fig:results_testset}). Precisely, the third (missing) coordinate of the gaze was calculated as intersection point between: i) an infinite ray starting from the camera center and passing through the gaze vector tip in the image plane at a distance of the focal length and ii) the virtual sphere having a radius of $10$ cm and centered in the centroid of the face keypoints (the same virtual sphere was used in the annotation phase to embed indirectly the z-axis in the gaze vector module, as described in Section~\ref{sec:gaze_regressor}). 

\begin{table*}[!h]
    \renewcommand{\arraystretch}{1.2}
    \caption{\textbf{End-to-end performance.} Mean and standard deviation of metrics used to evaluate the whole pipeline having the models trained with the data acquired from the iCub's camera or/and from the Intel RealSense camera and tested on the data collected solely from the iCub's eye.}
    \centering
    \begin{tabular}{|c|c|c|c|}
         \cline{2-4}
         \multicolumn{1}{c|}{} & \multicolumn{3}{|c|}{\textit{iCub Test set}}  \\
         \hline
         \textit{Train set} & \textit{iCub} & \textit{Realsense} & \textit{iCub + Realsense} \\
         \hline
         percentage classified as \textit{workspace} [\%] & $0.94 \pm 0.04$ & $0.76 \pm 0.08$ & $\mathbf{0.94 \pm 0.03}$ \\
         \hline
         2D Gaze RMSE & $5.68 \pm 0.77$ & $5.97 \pm 0.73$ & $\mathbf{5.69 \pm 0.68}$ \\
         \hline
         3D Gaze angular error [degree] & $13.11 \pm 1.25$ & $13.18 \pm 1.11$ & $\mathbf{12.80 \pm 1.12}$ \\
         \hline
    \end{tabular}
    \label{tab:end-to-end}
\end{table*}

From now on, we refer to the learning pipeline trained with the joined dataset coming from both iCub’s right eye and Realsense camera since it reported higher performance metrics. Figure~\ref{fig:results_testset} shows some sample frames to illustrate the input and the expected output from the pipeline in different use cases.

\subsection{Comparison with State-of-the-art method}
\label{sec:comparison_soa}
In this Section the gaze regressor is compared with two solutions proposed in~\cite{VTD} and in~\cite{3DGazeNet}. We compared only the gaze regressor because architectures that combine as output the mutual/averted gaze, gaze towards the partner and the regression of the gaze vector cannot be found in the current literature (this is one of the major contributions of our proposed work).
Chong et al. in~\cite{VTD} proposed a novel spatial reasoning architecture to improve the attention estimation by exploiting the interaction between the scene and the head features. The architecture is composed of a scene convolutional layer that is regulated by the head convolutional layer via an attention mechanism, such that the model focuses on the scene region that the head is oriented to. The proposed approach was benchmarked on the GazeFollow dataset~\cite{recasens2015} and compared with the solutions proposed in~\cite{chong2018,lian2018,zhao2020} reaching the highest performance. In what follows we refer to this learning model as VTD (visual target detection model).

The solution proposed in~\cite{3DGazeNet}, instead, aimed at developing a general gaze estimation model (named 3DGazeNet) that can be directly employed in novel environments without using domain adaptation techniques. Specifically, they faced the problem of gaze estimation considering the gaze output as regression of dense 3D eye meshes which was combined with the standard 3D vector regression. The final 3D gaze vector was calculated as mean direction of the two output components (eye meshes regression and gaze vector regression). Authors in~\cite{3DGazeNet} compared the proposed approach with other solutions for gaze generalisation existing in the literature (~\cite{bao2022,cheng2022b,wang2022}) testing them on two widely common gaze estimation datasets (MPIIFaceGaze~\cite{zhang2017} and GazeCapture~\cite{krafka2016}).

As our aim is to find the best performance in our experimental setup with the corresponding constrains, we run the VTD and 3DGazeNet algorithms on our test set collected from iCub camera. Specifically: 1) 3DGazeNet was used off-the-shelf since the authors provided the pre-trained model but the released code was incomplete and did not allow us to fine-tune the model on our train set; 2) VTD, instead, was both used off-the-shelf and fine-tuned on the collected dataset. Specifically, for a fair comparison, the VTD algorithm was fine-tuned using the same $5$ participant splits used to train our approach performing a warm training re-start with the pre-trained weights provided by the authors in~\cite{VTD}. The hyper-parameters were chosen empirically as follows: batch size of $16$, number of epochs equal to $20$ and a learning rate of $2.5e^{-4}$.
Furthermore, since the pretrained models give as output a gaze vector, which is normalized to the unit, all the algorithms' output used in the comparison were converted to gaze vectors with unitary module. The comparison is shown in Table \ref{tab:comparison}, where our method outperforms the others both in terms of RMSE and angular error.

\begin{table*}[!h]
    \renewcommand{\arraystretch}{1.2}
    \caption{\textbf{Comparison with state-of-the-art.} Our proposed approach is compared with 3DGazeNet~\cite{3DGazeNet} and VTD~\cite{VTD}. 3DGazeNet is used off-the-shelf, while VTD is used both off-of-shelf and after the fine-tuning. Reported metrics are the RMSE and the angular error between the ground-truth and the predicted gaze vector (lower is better).}
    \centering
    \begin{tabular}{|c|c|c|c|c|}
         \cline{2-5}
         \multicolumn{1}{c|}{} & \multicolumn{4}{|c|}{\textit{iCub Test set}}  \\
         \hline
         \textit{Method} & \textit{3DGazeNet}~\cite{3DGazeNet}  & \textit{VTD}~\cite{VTD} & \textit{VTD fine-tuned} & \textit{Ours} \\
         \hline
         Gaze RMSE & $0.38 \pm 0.03$ & $0.24 \pm 0.01$ & $0.18 \pm 0.02$ & $\mathbf{0.12 \pm 0.02}$ \\
         \hline
         Gaze Angular error [degree] & $32.98 \pm 2.86$ & $19.56 \pm 0.76$ & $14.72 \pm 1.95$ & $\mathbf{10.04 \pm 1.73}$ \\
         \hline
    \end{tabular}
    \label{tab:comparison}
\end{table*}

\section{Conclusions}
\label{sec:conclusions}
In this paper, we proposed a learning architecture for gaze estimation in face-to-face human-robot interaction in table-top scenarios. Specifically, our approach answers to three different questions: (i) is the human looking at the robot's eyes (mutual gaze)?, (ii) is the human looking at any robot's body part? and (iii) if the human is looking at the workspace, where is their gaze directed at? 
Additionally, we collected a novel dataset involving the humanoid robot iCub in order to cover the different gaze conditions addressed in our work.

We validated and compared the proposed approach with the state-of-the-art method proposed in~\cite{3DGazeNet} and ~\cite{VTD}, reporting a consistent improvement in all the considered performance metrics.

Major point of strength of our method is that it requires neither any additional hardware (e.g. external sensors, eye-tracking system) nor a robot with embedded high-quality and expensive eye-cameras (often not available in humanoid robots for space constraints). This system is specifically designed to support human-robot interaction experiments, in which the interaction between the robot and humans happens through objects placed on a table. This is a scenario widely adopted to study social cognition leveraging robots as tools for controlling the interaction between humans and other agents. Several studies in literature analyse gaze and gaze patterns as social cue that characterise human behaviour when interacting with others; better tools for estimating gaze can therefore contributing to better understanding human behaviour and affective processes in this context.     

Ongoing work focuses on extending the proposed gaze detection system combining it with other social cues (e.g. action recognition and natural language processing) in a multimodal architecture. The choice of using human keypoints as main features vector is aligned with this ongoing research direction because it allows extensions that leverage the information provided by the human pose estimation component (for example, gesture or action recognition).

Furthermore, our architecture poses a fundamental step to close the human-robot interaction loop where it is crucial increasing the robot awareness about the state and intention of the human partner, during human-robot collaborative tasks involving objects.

\begin{acks}
We thank Dr. Davide De Tommaso for his technical support in using the robot framework and set the laboratory for the data collection. Also, we thank Dr. Serena Marchesi for her support and assistance in recruiting the participants for the data collection. We also thank all the people who took part in the experiments.
\end{acks}

\section*{Ethics statement}
The studies involving human participants were reviewed and approved by the Comitato Etico Regione Liguria. The participants provided their written informed consent to participate in this study.

\section*{Data and code availability statement}
The anonymised data that support this study, the code and the learning trained models will be released upon acceptance. Further inquiries can be directed to the corresponding author.

\section*{Author contributions}
ML, EM, AW and LN conceived and discussed the project idea and the rationale. ML, EM and LN designed the learning architecture. ML collected the data, implemented the learning system, performed the experiments, analysed the data and deployed the algorithm on the iCub robot. ML, EM and LN discussed the results. ML wrote the manuscript. All authors revised the manuscript.

\section*{Funding}
This work received funding from the project Fit for Medical Robotics (Fit4MedRob) - PNRR MUR Cod. PNC0000007 - CUP: B53C22006960001 and Future Artificial Intelligence Research (FAIR) –
PNRR MUR Cod. PE0000013 - CUP: E63C22001940006.

\section*{Conflict of interest statement}
The authors declare that the research was conducted in the absence of any commercial or financial relationships that could be construed as a potential conflict of interest.

\section*{Supplementary material}
The Supplementary Material for this article can be found online at: [Link Supplementary Material].

\bibliographystyle{ACM-Reference-Format}
\bibliography{bibliography}

\end{document}


\title{Gaze estimation learning architecture as support to affective, social and cognitive studies in natural human-robot interaction \\ Supplementary Materials}

\author{Maria Lombardi}
\email{maria.lombardi1@iit.it}
\orcid{0000-0001-5792-5889}
\authornote{Corresponding authors.}
\affiliation{%
  \institution{Humanoid Sensing and Perception, Istituto Italiano di Tecnologia}
  \city{Genoa}
  \country{Italy}
}

\author{Elisa Maiettini}
\email{elisa.maiettini@iit.it}
\affiliation{%
  \institution{Humanoid Sensing and Perception, Istituto Italiano di Tecnologia}
  \city{Genoa}
  \country{Italy}
}

\author{Agnieszka Wykowska}
\email{agnieszka.wykowska@iit.it}
\affiliation{%
  \institution{Social Cognition in Human-Robot Interaction, Istituto Italiano di Tecnologia}
  \city{Genoa}
  \country{Italy}
}

\author{Lorenzo Natale}
\email{lorenzo.natale@iit.it}
\authornotemark[1]
\affiliation{%
  \institution{Humanoid Sensing and Perception, Istituto Italiano di Tecnologia}
  \city{Genoa}
  \country{Italy}
}

\renewcommand{\shortauthors}{Lombardi et al.}

\maketitle

The proposed approach resulted from a series of comparisons between different algorithms and configurations. Specifically for the multi-class classifier, the SVC was compared with the random forest algorithm and with a feed-forward neural network (Section~\ref{sec:comparison_multiclass}). For the sake of completeness in Section~\ref{sec:dataset_realsense}, we report also the performance of the system when trained and tested on the dataset acquired only from the Realsense camera.

\section{Multi-class classifier: Comparison between different algorithms}
\label{sec:comparison_multiclass}
In this Section, we make a comparison between the support vector classifier (SVC), the random forest algorithm and a feedforward neural network using as training samples the key-points from OpenPose. All the algorithms were trained and tested on the same train and test set (train set: $19$ out of $24$ participants, test set: $5$ participants). The train set was augmented as described in the main article.
Specifically, the hyper-parameters for the SVC with RBF kernel (parameters $C$ and $\gamma$) and for the random forest (number of estimators) were selected using an exhaustive search over a grid parameters. The parameters of the feed-forward neural network were tuned empirically, they are $3$ hidden layers fully connected with $100$ neurons each, learning rate set to $1e^{-3}$, ReLU as activation function and Adam optimiser. 
The performance, measured in terms of accuracy, precision, recall and F1 score are reported in Table \ref{tab:multiclass_classifier}. As in Section \textit{Results} of the main article, we considered mean and standard deviation on $k=5$ random splits of the dataset. The SVC performs better than the others for all metrics (values reported in bold in Table \ref{tab:multiclass_classifier}).

\begin{table}[h!]
    \renewcommand{\arraystretch}{1.1}
    \caption{\textbf{Multiclass classifiers comparison.} Mean and standard deviation of accuracy, precision, recall and F1 score are reported for the SVC, the random forest algorithm and the neural network trained with (a) the data acquired from the iCub's camera, (b) the Realsense camera and (c) both of them. All approaches are tested on the data collected only from the iCub's eye.}
    \centering
    \subcaption*{(a): Trained with the data acquired from the iCub's camera}
    \begin{tabular}{|c|c|c|c|}
         \hline
         Algorithm & SVC & Random Forest & Neural Network \\
         \hline
         Accuracy & $0.85 \pm 0.03$ & $0.84 \pm 0.02$ & $0.85 \pm 0.02$ \\
         \hline
         Precision & $0.75 \pm 0.05$ & $0.74 \pm 0.03$ & $0.74 \pm 0.03$ \\
         \hline
         Recall & $0.72 \pm 0.06$ & $0.67 \pm 0.06$ & $0.74 \pm 0.06$ \\
         \hline
         F1 score & $0.73 \pm 0.05$ & $0.69 \pm 0.05$ & $0.73 \pm 0.05$ \\
         \hline
    \end{tabular}
    \bigskip
    \subcaption*{(b): Trained with the data acquired from the Realsense camera}
    \begin{tabular}{|c|c|c|c|}
         \hline
         Algorithm & SVC & Random Forest & Neural Network \\
         \hline
         Accuracy & $0.76 \pm 0.05$ & $0.79 \pm 0.03$ & $0.80 \pm 0.04$ \\
         \hline
         Precision & $0.65 \pm 0.04$ & $0.64 \pm 0.03$ & $0.67 \pm 0.03$ \\
         \hline
         Recall & $0.69 \pm 0.04$ & $0.67 \pm 0.04$ & $0.69 \pm 0.05$ \\
         \hline
         F1 score & $0.66 \pm 0.04$ & $0.64 \pm 0.03$ & $0.67 \pm 0.04$ \\
         \hline
    \end{tabular}
    \bigskip
    \subcaption*{(c): Trained with the data acquired from the Realsense and iCub's camera}
    \begin{tabular}{|c|c|c|c|}
         \hline
         Algorithm & SVC & Random Forest & Neural Network \\
         \hline
         Accuracy & $\mathbf{0.87 \pm 0.03}$ & $0.82 \pm 0.02$ & $0.85 \pm 0.02$ \\
         \hline
         Precision & $\mathbf{0.79 \pm 0.06}$ & $0.67 \pm 0.04$ & $0.76 \pm 0.02$ \\
         \hline
         Recall & $\mathbf{0.75 \pm 0.05}$ & $0.64 \pm 0.06$ & $0.74 \pm 0.05$ \\
         \hline
         F1 score & $\mathbf{0.75 \pm 0.05}$ & $0.65 \pm 0.04$ & $0.75 \pm 0.03$ \\
         \hline
    \end{tabular}
    \label{tab:multiclass_classifier}
\end{table}

\newpage

\section{Performance of the system on the dataset collected from the Realsense camera}
\label{sec:dataset_realsense}
In table \ref{tab:dataset_realsense}, we report the performance for the SVC and the gaze regressor when both the train set and the test set were collected from the Realsense camera.

\begin{table}[h!]
    \renewcommand{\arraystretch}{1.1}
    \caption{\textbf{Performance reported with the dataset collected from the Realsense camera.} Mean and standard deviation over the $5$ splits of the dataset are reported for the metrics used to evaluate both the multiclass classifier and the gaze regressor. Both trainset and testset are collected from the Realsense camera.}
    \centering
    \begin{tabular}{|c|c|c|c|c|}
         \cline{2-5}
         \multicolumn{1}{c|}{} & \multicolumn{3}{|c|}{Multiclass classifier} & Gaze regressor \\
         \hline
         \hline
         Algorithm & SVC & Random Forest & NN Classifier & NN Regressor \\
         \hline
         Accuracy & $0.90 \pm 0.02$ & $0.87 \pm 0.03$ & $0.89 \pm 0.03$ & $--$ \\
         \hline
         Precision & $0.83 \pm 0.05$ & $0.81 \pm 0.07$ & $0.82 \pm 0.05$ & $--$ \\
         \hline
         Recall & $0.81 \pm 0.05$ & $0.74 \pm 0.05$ & $0.79 \pm 0.06$ & $--$ \\
         \hline
         F1 score & $0.81 \pm 0.05$ & $0.75 \pm 0.05$ & $0.80 \pm 0.06$ & $--$ \\
         \hline
         RMSE & $--$ & $--$ & $--$ & $5.86 \pm 0.90$ \\
         \hline
    \end{tabular}
\label{tab:dataset_realsense}
\end{table}
